# INDEXER BASED DYNAMIC WEB SERVICES DISCOVERY

Saba Bashir, M.Younus Javed,Farhan Hassan Khan
National University of Science & Technology
Rawalpindi, Pakistan

Aihab Khan, Malik Sikandar Hayat Khiyal
Fatima Jinnah Women University
Rawalpindi, Pakistan

*Abstract*-**Recent advancement in web services plays an important role in business to business and business to consumer interaction. Discovery mechanism is not only used to find a suitable service but also provides collaboration between service providers and consumers by using standard protocols. A static web service discovery mechanism is not only time consuming but requires continuous human interaction. This paper proposed an efficient dynamic web services discovery mechanism that can locate relevant and updated web services from service registries and repositories with timestamp based on indexing value and categorization for faster and efficient discovery of service. The proposed prototype focuses on quality of service issues and introduces concept of local cache, categorization of services, indexing mechanism, CSP (Constraint Satisfaction Problem) solver, aging and usage of translator. Performance of proposed framework is evaluated by implementing the algorithm and correctness of our method is shown. The results of proposed framework shows greater performance and accuracy in dynamic discovery mechanism of web services resolving the existing issues of flexibility, scalability, based on quality of service, and discovers updated and most relevant services with ease of usage.**

## I. INTRODUCTION

As an enabling technology, web services are software components that are used to present services on internet. Each web service has standard interface through which it communicates with other web services in order to provide different functionalities. Service providers register their web services in registries, typically in UDDI (Universal Description Discovery and Integration). HTTP (Hyper Text Transfer Protocol) protocol is used for communication of SOAP (Simple Object Access Protocol) messages exchanged between service providers and consumers.

With the increasing technology it is becoming more critical to find a service that satisfies user requirements. Service discovery is a mechanism used to find the appropriate service that meets desired criteria. Three main roles in web services are providers, consumers and registries. Service providers publish their web services in registries and clients discover their desired services by querying them using SOAP message protocol [1].

In manual discovery process a human uses a discovery service, which involves the input of user requirements and through matchmaking process, output the desired results [2]. The main obstacle in this mechanism is heterogeneity problem [3]. Different types of heterogeneities are:

- Technological heterogeneities (different platforms or different data formats)
- Ontological heterogeneity (difference in domain specific terms and concepts)
- Pragmatic heterogeneity (different development of domain-specific processes)

This makes the technology limited in terms of efficiency and availability. Whereas in dynamic discovery process an agent performs the above task either at design or run time, but there are different constraints of this process for example interface requirements, standardization of protocols and trust of stakeholders etc. This paper focuses on the dynamic web services discovery approach.

With the increasing number of web services, it is becoming more difficult to manage description on distributed environment such as on web. Key limitations of current approaches for dynamic web services discovery are:

1. Lack of fully dynamic web services discovery mechanism.
2. Lack of mechanism that solves heterogeneity problems.
3. Clients cannot learn from past experiences to improve future decisions.
4. Need of mechanism which store and retrieve up to date information on consumer's request.
5. Need of mechanism which gives in time response of query in case of searching services from web.

6. In existing approaches, clients have to search from web (registries) each time when they request for services which is time consuming task and requires lot of effort.

7. Service providers and consumers don't use same language for registration and discovery.

Keyword based discovery is one possible way to discover the web services. The advantage of UDDI is that it filters and ranks services quickly. The drawback is that it contains only metadata. It is based on keyword matching and it has only single search criteria. In some cases it may not produce the desired results due to the words that are semantically related with each other [2].

In order to take a full advantage of web services, user must discover and invoke them dynamically. Semantic web services are a possible solution for automated web service discovery in which services are semantically described and accessed. It resolves the heterogeneity as well as interoperability problems. Ontologies are used for manipulation of data. There are two drawbacks of semantic web services discovery approach. First, it is impossible for all service providers to publish their services in same ontology. Second, already existing web services does not have associated semantics and it is not possible to convert all of them into same ontology [4].

*A. Contribution*

This research paper proposed a framework that presents solutions to most of the problems in dynamic web service discovery. More precisely, it uses the translator to convert service requests and responses in one form that can be easily used by the system. Indexer is used to index services to simplify search process. CSP solver is used to select most appropriate service. Aging factor helps to maintain updated information. The discovery process is fastened by the use of local cache which allows clients to discover services based on previous experiences.

The organization of paper is such that Section 2 describes the previous research on dynamic services discovery. In section 3, some basic information about service discovery and dynamic service discovery is given. Section 4 presents the detailed overview of proposed technique including framework and key mechanisms. Section 5 describes the working and algorithms of proposed framework. Section 6 presents the implementation and analysis of proposed framework. Finally, the conclusion and future work is given in section 7.

## II. RELATED WORK

Web services are XML (Extensible Markup Language) based software components [5, 6]. These components can be retrieved based on signature matching and the discovery process is based on interface matching. The advantage of this approach is that discovery process totally depends on properties of component. WSDL (Web Services Description Language) is used as interface definition language to describe the functionality of web services. It is an XML based format used to describe both abstraction operations and network bindings [7].

UDDI is used in discovery process where keywords are matched by using intelligent algorithms. The research is XML schema matching where various string comparison techniques are applied like suffix, prefix, and infix for keyword matching. This technique is useful where many acronym, jargons etc are used in web services [8].

Liang-Jie Zhang, Qun Zhou [9] proposed a dynamic web services discovery framework for web services representation chain. It solves the manual search process problems in linked documents. The multiple linked documents are searched using WSIL (Web Services Inspection Language) chains and results are aggregated and returned them to users all at once. It also re-explores the nested WSIL documents by using pre-fetched link calculation and caching methods.

Sudhir Agarwal [10] proposed a goal specification language to discover the web services. It describes that the constraints should be specified on functional and non functional properties in goal specification language. Then this goal specification language is used to discover the desired services. The language is novel combination of SHIQ(D) and $\mu$-calculus. It covers resources, behavior and policies of agents involve in discovery process. Java API (Application Programming Interface) is used to describe goals and concrete syntax is used so that end users can easily write the formulas through interface.

Fu Zhi Zhang et al. [2] presented an algorithm for dynamic web services discovery. The proposed algorithm is based on composition of web services with the process model. It is based on OWL-S (Ontology Web Language-Semantic) ontology. Unlike previous proposed models for dynamic discovery it can have multiple executions and matchmaking for a single service request and return desired service.

Paul Palathingal [11] presented an agent based approach for service discovery and utilization. Agents acts instead of users and dynamically discover, invoke and execute web services. Through this technology sender object must not know the receiver's address. Also the interoperability between distributed services is being achieved through these software components. The selected agents match the desired query from their respected repositories and send the retrieved results to composition agent who composes the web services and sends back to requester.

Stephan Pöhlsen and Christian Werner [12] proposed a dynamic web services discovery framework in large networks. It solves the discovery problem of mobile ad-hoc networks by introducing a Discovery Proxy (DP). Through Discovery proxy clients can easily search desired web services through service providers and get updated information as services does not have to register in proxy unlike UDDI. The drawback of given technique is that all information is not centralized and global DNS (Domain Name System) cannot be easily located.

Aabhas V. Paliwal [13] proposed dynamic web services discovery through hyper clique pattern discovery and semantic association ranking. It solves the discovery problem that is, when user input a query, multiple services that meet the desired criteria are retrieved as output and user cannot identify most relevant service. The proposed model presents the mechanism that selects the most relevant web service from all outputs. The approach is a combination of semantic and statistical metrics.

A number of web services algorithm use service profile method for dynamic web services discovery. [16] proposed the discovery mechanism based on web services composition and service profile. The proposed algorithm does not describe the internal behavior for web services composition.

Although a number of dynamic web services techniques have been introduced, there is a need of dynamic services discovery mechanism which eliminates the problems in current discovery approaches. This paper presents an approach towards these solutions and proposes a framework that is efficient, reliable, scalable, flexible and fault tolerant.

## III. PRELIMINARIES

This section describes the web services discovery mechanism and differentiates between different approaches used for service discovery.

### A. Web Services Discovery

Web services discovery is the process of finding the appropriate services on the web that meets the user's desired requirements. All Service providers register their web services in registry called UDDI. Service requesters search the appropriate service providers available on the registry. Each service that is registered on UDDI has WSDL. WSDL is a description language which describes complete detail about web services for example its methods, protocols and ports etc. Web service interface is described in WSDL and used by the client to discover and invoke web services. SOAP message is sent to server asking for certain service and it replies with SOAP response message. Web service discovery and invocation process is shown in Fig 1. [14]

### B. Web Service Discovery Approaches

Different approaches for discovery mechanism as discussed in [15] are:

1. UDDI: It is a standard for web services registration, discovery and integration.

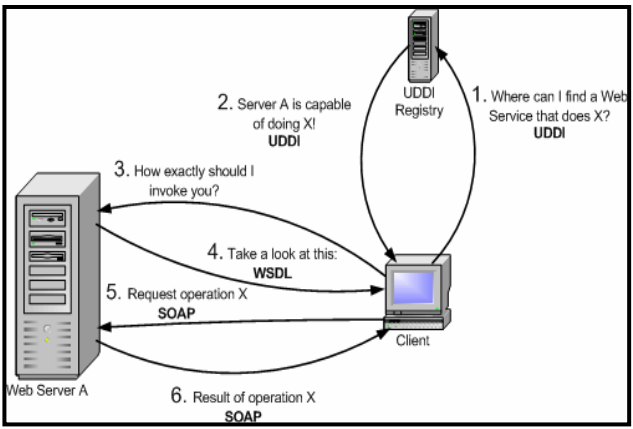

Fig 1: Web services discovery and invocation process

2. Service directories/portals: They provide the web services using focused crawlers and provide searching through HTML (Hyper Text Markup Language) interface.
3. Web Search Engines: They are most widely used in today's web. They search the web services by retrieving information from WSDL descriptions.

These approaches can be compared by focusing on following two aspects: 1) Number of services, 2) Quality of information associated with them. It is concluded from this research that browsers can give the desired results but one have to spend too much time for filtering the desired service which requires a lot of effort and it is a time consuming task. Only dedicated portal pages produce best results.

## IV. PROPOSED FRAMEWORK

Proposed framework of dynamic services discovery mechanism is shown in Fig 2. Web services discovery framework includes the following components:

1. *Translator:* Most service discovery frameworks distinguish between external and internal specification languages. Translator is used for translation of user input to a form that can be efficiently used by the system. It is also used to translate the system response to a form that can be easily understandable for the user.
2. *Matching Engine:* Key component that is used for dynamic discovery mechanism. It matches the input to registered services using keyword and fuzzy logic based

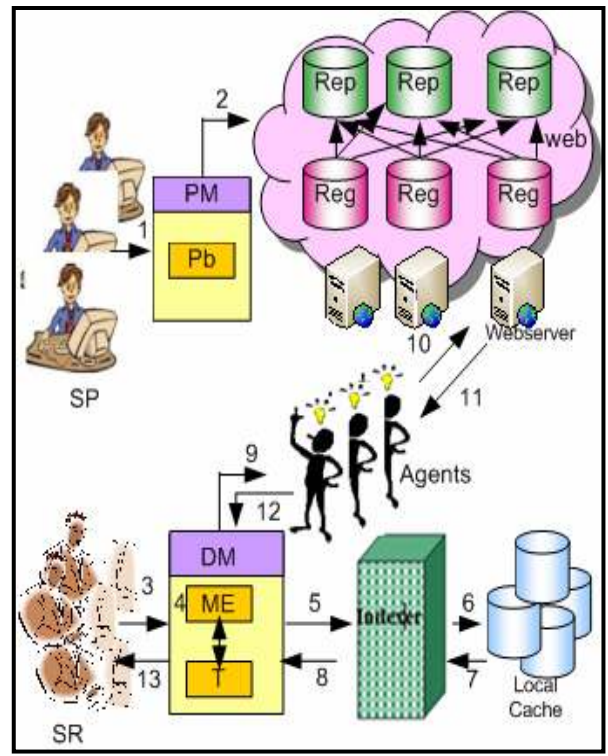

Fig 2: Proposed framework of dynamic web services discovery

The abbreviations used in framework diagram are:

SP= Service Provider      Reg= Registry
SR= Service Requester     Rep= Repository
DM= Discovery Module     T= Translator
PM= Publisher Module     ME=Matching Engine

discovery. Also it stores references of searched services in repositories based on classification and indexing. Advantage of keyword based discovery is that it filters and ranks the services quickly. It also uses CSP solver to filter services based on specified constraint e.g. business name, service name etc.

3. *Indexer:* Indexing mechanism is used to enhance the searching capability. Indexer is used to index services stored in local cache. In indexing mechanism first the index structure is created and then this structure is queried and results are returned. Indexing mechanism is shown in Figure 3.

4. *Local Cache:* Local cache concept is introduced so that users can learn from past experiences. Local cache is used to store the web services. It consists of categorized repositories. Web services that are discovered by matching engine are returned to the requester. It also stores the service reference in local cache using indexing mechanism for future reference. A timestamp is associated with each service reference present in the local cache to help maintain up to date cache.

5. *Registry:* It is authorized, centrally controlled store of web services. Multiple service providers use multiple registries to publish their services in registries on global web.

6. *Publisher:* Publisher is used to publish the web services in registries. Through publisher web service providers can save, update and delete their web services.

7. Agent: Agent is used to search web service from its own listed registries. It sends the multiple threads to each registry, and returned the discovered services to matching engine.

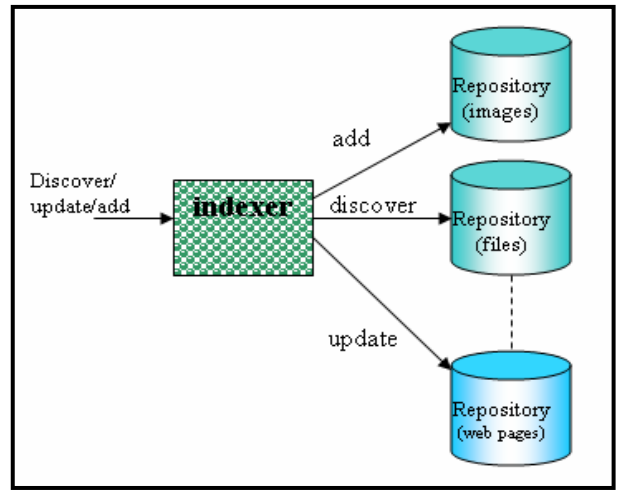

Fig 3: Indexing mechanism

*Key Mechanisms*

Following key mechanisms are used in proposed framework to enhance the discovery mechanism.

1. *Classification of Services:* Web services are classified or categorized into different categories in local cache. The classification is done based on different data types for example images, files, html docs etc. For each category different database is maintained with timestamp. This mechanism introduces fast and efficient retrieval of web services.

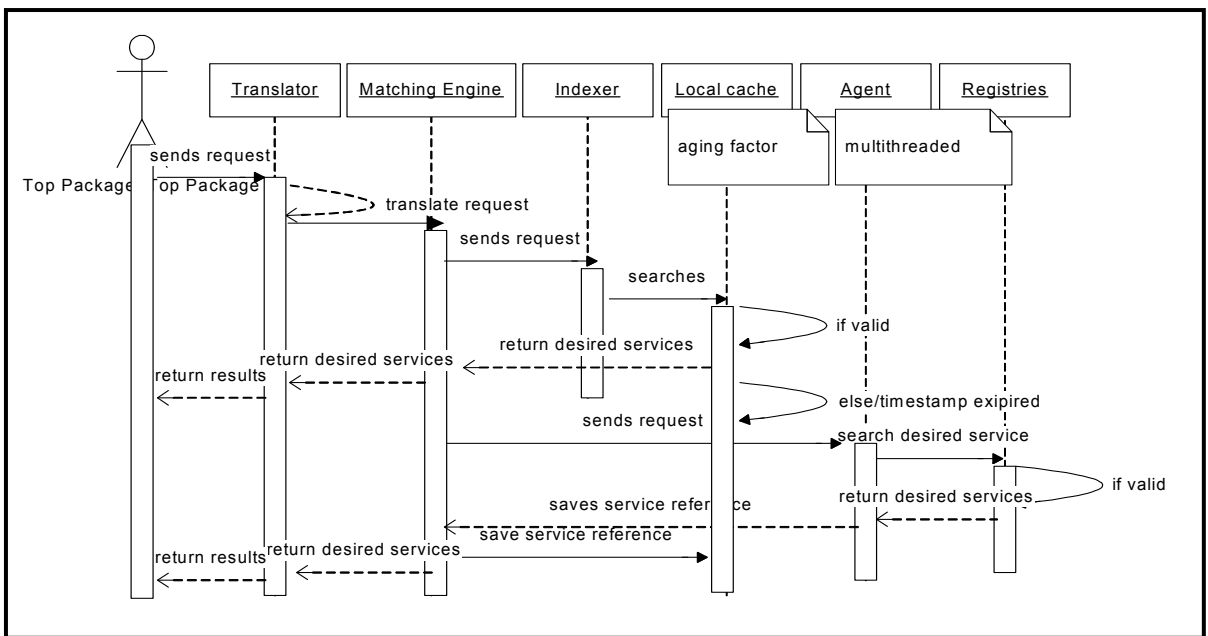

Fig 4: Sequence diagram of proposed framework

2. *Multithreading:* The agent is based on multithreading and opens thread for each registry listed with itself. Multiple agents have been used in our proposed framework and each agent will search desired service from its listed registries and returned the results to matching engine.

## V. PROPOSED TECHNIQUE

*A. Working.*
The detailed flow of proposed discovery mechanism is explained as follows:

1. Service providers register their web services in registries on global web through registration module.
2. Service requester requests for web service through discovery module.
3. Translator is used for translation of user input to a form that can be efficiently used by the system. From translator request goes to matching engine.
4. Matching engine tries to discover the desired service from local cache and passes request to indexer. Indexer searches the desired service from local cache based on indexing and categorization. If the results are valid then indexer returns desired results to matching engine.
5. Otherwise matching engine tries to discover the desired service from registries available on the web, using multithreaded agent approach. Agents search the services from its listed registries and send the discovered services to matching engine.
6. A timestamp is maintained with each service reference which has a specific value. And web services are updated in local cache from web when timestamp expires.
7. Matching engine stores the service references in local cache for future reference. Also it sends result to translator which translates result into user understandable form and results are returned to requester.

*B. Pseudo Code:*
The pseudo code of proposed technique is given as:

---

**Algorithm**: Dynamic Web Services Discovery
**Input:** Request for Web service
**Output:** Desired Service

Services registered in UDDI registries;
User enters input request for web service;
for each input
 Translator translates input;
 Input goes to Matching Engine;
 Matching Engine passes request to
 Indexer;
 Indexer search from local cache
 based on classification and indexing;
 Indexer search from local cache
 based on classification and indexing;

---

If service found in local cache
 Return result to user;
If no service found or timestamp expires
 Matching Engine passes request to
 Agents;
 Agents discover service from its listed
 UDDI registries;
 Keyword and fuzzy logic based
 matching from registries;
 CSP solver filters matched services
 based on specified constraints;
 Matching Engine add references of
 matched services in local cache through
 indexing;
 Service references added in local cache
 for future reference;
 Return results to user through translator;
else
 Return no service found;

---

## VI. IMPLEMENTATION & EVALUATION

The implementation has been done using Netbeans 6.5 and open source UDDI which is Apache jUDDIv3 (java implementation of Universal Description Discovery and Integration version 3). Ruddi is used to access UDDI registry which is java API for developing, deploying reliable, scalable and secure application that query and publish to UDDI registries. UDDI4J and JAXR (Java API for XML Registries) is also java API for accessing registries but the drawbacks are that UDDI4J only supports jUDDI version 2 and JAXR is for all XML based registries for example UDDI, ebXML etc. Xerces APIs are also used for parsing XML request and responses. We have used apache jUDDI version3 and multiple jUDDI's on multiple servers by implementing multithreaded approach.

For testing we have published multiple businesses and services on registry. We have used WordNet for synonyms and Rita-Wordnet is a java API to access WordNet. WordNet is used for matching businesses and services from registry. Users can search businesses and services by synonyms which are already saved in WordNet. The implementation mainly comprises of UDDI registries, matching engine, service providers, service requesters and dynamic services discovery mechanism. Fig 5 shows the GUI of our dynamic web services discovery application. User sends discovery request using our application which then goes to the translator. Translator send request to matching engine which tries to discover the desired service from local cache based on indexing. As this service is not present in local cache, it sends request back to Matching Engine. Matching Engine sends request to agent. Agent sends multiple threads to own multiple listed registries and starts searching services. It discovers the matched services and returns the valid services to Matching Engine. Matching

Request body:
<find_business generic="2.0"
xmlns="urn:uddi-org:api_v2">
<findQualifiers/>
<name>Microsoft</name>
</find_business>

Response body:
<businessList generic="2.0"
operator="ms.com" truncated="false"
xmlns="urn:uddi org:api_v2">
<businessInfos>
<businessInfo
businessKey="c13cc7b2-642d-41d0-
b2dd-7bb531a18997">
<name xml:lang="en">Microsoft DRMS
Dev</name>
<serviceInfos>
<serviceInfo businessKey="c13cc7b2
642d-41d0-b2dd-7bb531a18997"
serviceKey="6166f8b2-436d-4001-9f68-
f37ff8b47ea3">
<name
xml:lang="en">Certification</name>
</serviceInfo>
<serviceInfo businessKey="c13cc7b2-
642d-41d0-b2dd-7bb531a18997"
serviceKey="7ae6c133-4471-4deb-93a5-
1158aaa826b8">
<name xml:lang="en">Machine
Activation</name>
</serviceInfo>
<serviceInfo
businessKey="c13cc7b2-642d-41d0-
b2dd-7bb531a18997"
serviceKey="52616482-653c-45f3-
ae08-e4d4ca8b66c2">
<name xml:lang="en">Server
Enrollment</name>
</serviceInfo>
</business nfo>
....
</businessInfos>
</businessList>

Fig 5: XML based request response message

Engine sends results to user by using translator, and also stores service references of these services in local cache with timestamp. Next time when any user searches for the same service then results will be found from local cache. If timestamp expires for this cached service, then Matching Engine will search query from web UDDI registry. XML based request and response messages for service discovery are shown in Figure 6.

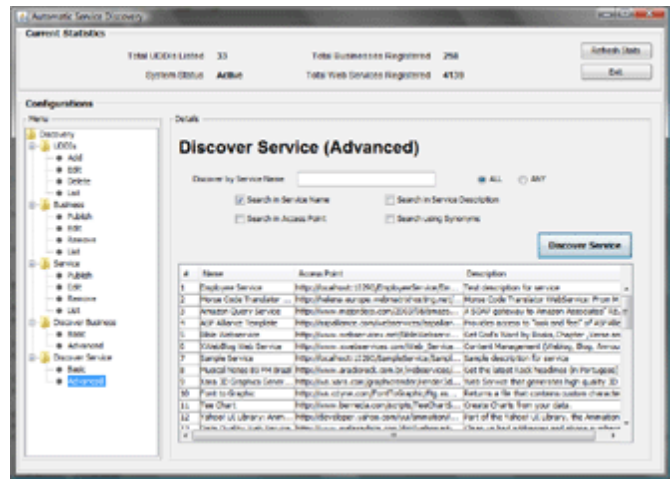

Fig 6: GUI of proposed discovery Application

### A. Experimental Evaluation and Analysis

Our algorithm accurately discovers the web services for requesters. The effectiveness of algorithm is proved by calculating their precision and discovery time.

Precision is the proportion of services that meets user's requests from all the discovered services.

Our algorithm has better performance and improved the quality of results. Figure 7, 8 show the efficiency and precision of proposed algorithm as compared to [2] and [16].

We have taken multiple number of web services and calculate their discovery time. Figure 7 shows the graph of number of services against discovery time and Figure 8 shows the graph of number of discovered services against precision.

In the proposed technique, the local cache technique improves the time taken for dynamic services discovery. Also the timestamp approach helps to make use of any new services that are available on the registry.

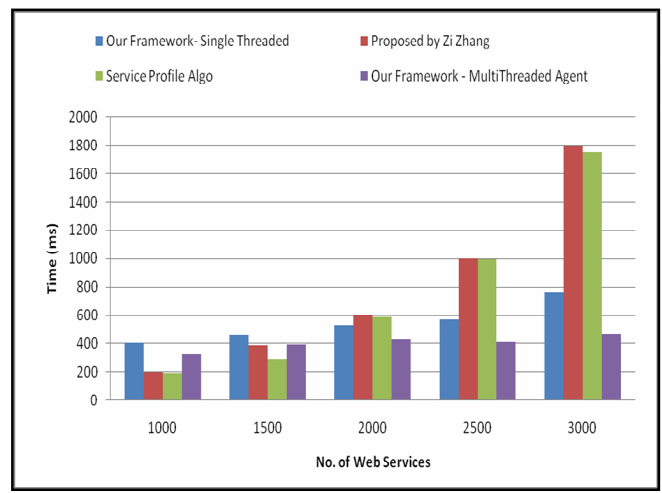

Fig 7: The Efficiency of proposed algorithm

In Fig 7, Single threaded and multithreaded concept is introduced. In Single threaded approach, Matching Engine discovers services from web, based on single search criteria whereas in multithreaded approach, multithreaded agents are used to discover services.

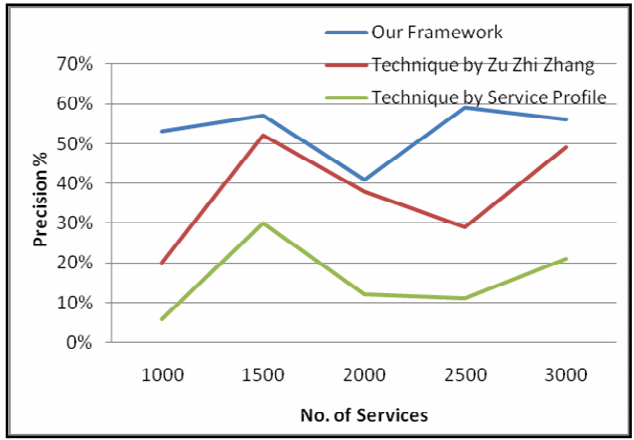

Fig 8: The precision of proposed algorithm

## VII. CONCLUSION & FUTURE WORK

This paper presents a framework for automatic, dynamic web services discovery and utilization. The framework is flexible, scalable and new services can easily be updated in local cache.

Through the proposed approach requester always retrieve up to date services because a timestamp is attached with each service reference in local cache and when it expires, services are updated. Also through local cache there is fast retrieval of services as requester does not need to search the web each time for discovery of services. If they are present in local cache then services can easily be discovered in less time. Interoperability between service providers and requesters is achieved through Translator.

CSP solver selects the service satisfied specified constraints, which is a part of Matching Engine. Thus the proposed algorithm fix current issues of dynamic web services discovery. In future, the framework can be extended by adding intelligent service aggregation algorithms based on AI planning for discovery.